\documentclass[final,5p,times,twocolumn,nopreprintline]{elsarticle}

\usepackage{framed,multirow}

\usepackage{amsmath}
\usepackage{amssymb}

\usepackage{placeins}
\usepackage{booktabs,arydshln}
\usepackage{subfig}
\usepackage{lipsum} 
\usepackage{array}
\newcolumntype{P}[1]{>{\centering\arraybackslash}p{#1}}
\usepackage{tabularx} 
\usepackage{bm}
\usepackage{siunitx}

\usepackage{pgfplots}
\usepackage{tikz}
\usetikzlibrary{positioning}
\usetikzlibrary{arrows.meta}
\usetikzlibrary{calc}
\usetikzlibrary{shapes.geometric}
\usepackage{xcolor}

\usepackage{pifont}
\usepackage{suffix}		
\newcommand\sref[1]{\protect\subref{#1}}
\WithSuffix\newcommand\sref*[1]{\protect\subref*{#1}}

\usepackage{hyperref}
\usepackage{url}

\usepackage{color, colortbl}
\definecolor{Gray}{gray}{0.9}
\newcolumntype{g}{>{\columncolor{Gray}}c}

\makeatletter
\def\adl@drawiv#1#2#3{%
        \hskip.5\tabcolsep
        \xleaders#3{#2.5\@tempdimb #1{1}#2.5\@tempdimb}%
                #2\z@ plus1fil minus1fil\relax
        \hskip.5\tabcolsep}
\newcommand{\cdashlinelr}[1]{%
  \noalign{\vskip\aboverulesep
           \global\let\@dashdrawstore\adl@draw
           \global\let\adl@draw\adl@drawiv}
  \cdashline{#1}
  \noalign{\global\let\adl@draw\@dashdrawstore
           \vskip\belowrulesep}}
\makeatother

\usepackage{fancyhdr}
\begin{document}

\begin{frontmatter}

\title{Motion-Corrected Moving Average: Including Post-Hoc Temporal\\Information for Improved Video Segmentation}

\author[1]{Robert Mendel\corref{cor1}}
    \cortext[cor1]{
    Corresponding authors. Both authors contributed equally. \\
    \indent \indent \hspace{-1.2mm} \textit{E-mail addresses:}\\ \href{mailto:robert1.mendel@oth-regensburg.de}{robert1.mendel@oth-regensburg.de}; \href{mailto:tobias.rueckert@oth-regensburg.de}{tobias.rueckert@oth-regensburg.de} \\
	This article is published under the CC BY-NC-ND license \\(\url{https://creativecommons.org/licenses/by-nc-nd/4.0/}).
        }
    \author[1]{Tobias Rueckert\corref{cor1}}
    \author[2]{Dirk Wilhelm}
    \author[3,4]{Daniel Rueckert}
    \author[1,5]{Christoph Palm}
    
    \address[1]{Regensburg Medical Image Computing (ReMIC), Ostbayerische Technische Hochschule Regensburg (OTH Regensburg), Regensburg, Germany}
    \address[2]{Department of Surgery, Faculty of Medicine, Klinikum rechts der Isar, Technical University of Munich, Munich, Germany}
    \address[3]{Artificial Intelligence in Healthcare and Medicine, Klinikum rechts der Isar, Technical University of Munich, Munich, Germany}
    \address[4]{Department of Computing, Imperial College London, London, UK}
    \address[5]{Regensburg Center of Health Sciences and Technology (RCHST), OTH Regensburg, Regensburg, Germany}

\begin{abstract}
Real-time computational speed and a high degree of precision are requirements for computer-assisted interventions.
Applying a segmentation network to a medical video processing task can introduce significant inter-frame prediction noise.
Existing approaches can reduce inconsistencies by including temporal information but often impose requirements on the architecture or dataset. This paper proposes a method to include temporal information in any segmentation model and, thus, a technique to improve video segmentation performance without alterations during training or additional labeling.
With Motion-Corrected Moving Average, we refine the exponential moving average between the current and previous predictions.
Using optical flow to estimate the movement between consecutive frames, we can shift the prior term in the moving-average calculation to align with the geometry of the current frame.
The optical flow calculation does not require the output of the model and can therefore be performed in parallel, leading to no significant runtime penalty for our approach.
We evaluate our approach on two publicly available segmentation datasets and two proprietary endoscopic datasets and show improvements over a baseline approach.
\end{abstract}

 
\end{frontmatter}

\section{Introduction}
The topic of real-time video segmentation has widespread applicability for computer-aided interventions \cite{garcia2017real,lin2021multi,wang2021efficient}.
However, including temporal information in the architectural design of the segmentation model can adversely affect the runtime performance. If a video segmentation model operates on the concatenation of inputs or features \cite{videotransformer,vps}, the computational load per output increases.
In addition to the runtime overhead and the more time- and hardware-intensive training, this can also affect data acquisition.
In the medical domain, where labeling cannot be easily outsourced, a requirement of fully labeled image sequences to enable temporal learning is difficult to realize.
Even just requiring videos for training, although unlabeled, is a limitation that has to be considered.

A low-overhead approach to include temporal information is to compute an exponential moving average (EMA) between the current and past predictions.
The exponential moving average has widespread applications in many deep learning tasks \cite{Vop2018,moco,ecmt} and can help reduce the inter-frame variability for video segmentation. 
However, the larger the emphasis on the past term of the EMA, the greater the discrepancy between the result and the current scene.
Therefore, determining the weighting factor in the EMA calculation is a choice between outlier reduction and shape-accurate segmentation.

Optical flow algorithms, more recently implemented with neural networks \cite{flownetv2} or classically computed from image features \cite{farne,siftflow}, have found applications in video segmentation and object detection tasks.
Warping the current prediction or latent state of the model with an estimated flow field has been proposed in two ways:
\begin{itemize}
    \item Warping from a keyframe state to intermediate results between the current and the next keyframe to skip the full forward pass \cite{warptoskipobj,warptoskipseg}.
    \item Taking a range of neighboring frames, estimating the motion to a keyframe, and aggregating the results for a single prediction \cite{flowguidedobject,medicalflowguidedseg}.
\end{itemize}
While the first approach is targeted at video processing, it forfeits accuracy on the intermediary frames. The second approach can increase the accuracy but requires warping and flow estimation for all supporting images, which introduces a considerable load during inference.

In this paper, we propose to combine optical flow with EMA, using the estimated pixel displacement to adjust the past term in the exponential moving average calculation.
Moving the computations into the feature space enables temporal information during inference without special training requirements or a complex architecture.
The procedure is light on the hardware, as both the optical flow and the feature-encoding forward pass can be run in parallel.
Continuously updating the feature representation offers the positive effects of EMA without the undesirable ghosting caused by overemphasizing the past term.
We demonstrate these effects on two proprietary and two public datasets and show how the effects prevail on subsets with large or small amounts of motion.

\section{Related Work}

The following presents an overview of related research organized by developments regarding image semantic segmentation, video semantic segmentation, and video semantic segmentation related to medical tasks.

\subsection{Image Semantic Segmentation}

Single-frame segmentation based on deep-learning approaches has achieved remarkable results in recent years and forms the basis for many video semantic segmentation methods \cite{fayyaz2016stfcn, gadde2017semantic, nilsson2018semantic, hur2016joint, chandra2018deep}.
The introduction of the concept of Fully Convolutional Networks (FCN) \cite{Shelhamer2017FullyCN} by replacing fully-connected layers through convolutional operations provided a significant contribution to this and laid the foundation for the development of high-quality segmentation methods.
Often, segmentation architectures consist of an encoder part extracting semantically relevant features by downsampling from an input image and a decoder part performing the reconstruction of the spatial context.
Based on these encoder-decoder-based architectures and the FCN networks, methods are developed to improve accuracy, for example, by using dilated (atrous) layers in the network architecture \cite{Chen2014SemanticIS, Chen2016DeepLabSI,Chen2017RethinkingAC,deeplab,Yu2015MultiScaleCA} or by further refining the network results using fully-connected condition random fields (CRF) \cite{Chen2014SemanticIS,Chen2016DeepLabSI}.
Using advanced backbone architectures in the encoder \cite{resnet,Huang2016DenselyCC,Howard2017MobileNetsEC} also leads to higher quality results \cite{resnet,Huang2016DenselyCC} or to faster computation time \cite{Howard2017MobileNetsEC}.
To capture multi-scale contextual information, architectures such as PSPNet \cite{Zhao2016PyramidSP} and DeepLab \cite{Chen2016DeepLabSI,deeplab} use spatial pyramid pooling \cite{He2014SpatialPP} and atrous spatial pyramid pooling \cite{Chen2016DeepLabSI,Chen2017RethinkingAC,deeplab}.

\subsection{Video Semantic Segmentation}

Video semantic segmentation approaches incorporate spatial and temporal context and can be divided into two categories.
The primary goal of methods in the first category is to increase segmentation accuracy based on the temporal information obtained using unlabeled intermediate frames.
For this purpose, many methods apply existing segmentation networks frame-by-frame and try to achieve better results by architecture enhancements as well as by adding additional modules \cite{fayyaz2016stfcn, gadde2017semantic} or by propagating the temporal information within the network structure \cite{nilsson2018semantic, hur2016joint, chandra2018deep}.
In \cite{fayyaz2016stfcn}, the authors present a module based on a long short-term memory (LSTM) architecture for capturing the temporal dependencies in successive video frames.
The work of \cite{gadde2017semantic} warps features of previous frames with the optical flow and processes them together with the features of the current image to obtain the final prediction.
Another idea is to use gated recurrent units to propagate semantic labels to non-annotated intermediate frames \cite{nilsson2018semantic}.
Other work jointly predicts optical flow and temporally consistent semantic segmentation with the idea that the two tasks leverage each other \cite{hur2016joint}.
The method described in \cite{chandra2018deep} focuses on the joint training of a convolutional neural network (CNN) for single image segmentation and a CRF to account for the temporal component.
However, using additional feature aggregation modules on top of existing segmentation networks or within network architectures is associated with additional runtime costs at inference time, which is difficult to reconcile with the necessary real-time capability of methods.

The second category of approaches attempts to reduce the computational cost by leveraging temporal information, usually by employing already computed features of previous frames \cite{shelhamer2016clockwork, Zhu2016DeepFF, Mahasseni2017BudgetAwareDS, Xu2018DynamicVS, Hu2020TemporallyDN}.
This can be achieved by using a multi-stage FCN, which reuses features at different stages from previous frames to reduce computation time \cite{shelhamer2016clockwork}.
Other works perform the elaborate segmentation only on sparse key frames and transfer the deep feature maps to intervening video frames using optical flow feature warping \cite{Zhu2016DeepFF}.
Optimal keyframe selection and propagation of CNN-based predictions to unlabeled neighboring frames is another approach, where propagation of runtime-intensive segmentations can be realized either by optical flow-based mapping or a very lightweight network \cite{Mahasseni2017BudgetAwareDS} or by dynamically discriminating between a CNN architecture and a flow network for different regions within a frame with subsequent feature warping \cite{Xu2018DynamicVS}.
In \cite{Hu2020TemporallyDN}, the authors present an approach based on the fact that high-level information in a deep CNN can be approximated by shallow features, which they obtain by distributing feature extraction over several successive frames by sub-networks with shared weights and then aggregating these features by an attention propagation module.
However, the focus on reducing computational cost often results in lower segmentation accuracy.
\begin{figure*}[t]
\centering
\includegraphics[width=0.85\textwidth]{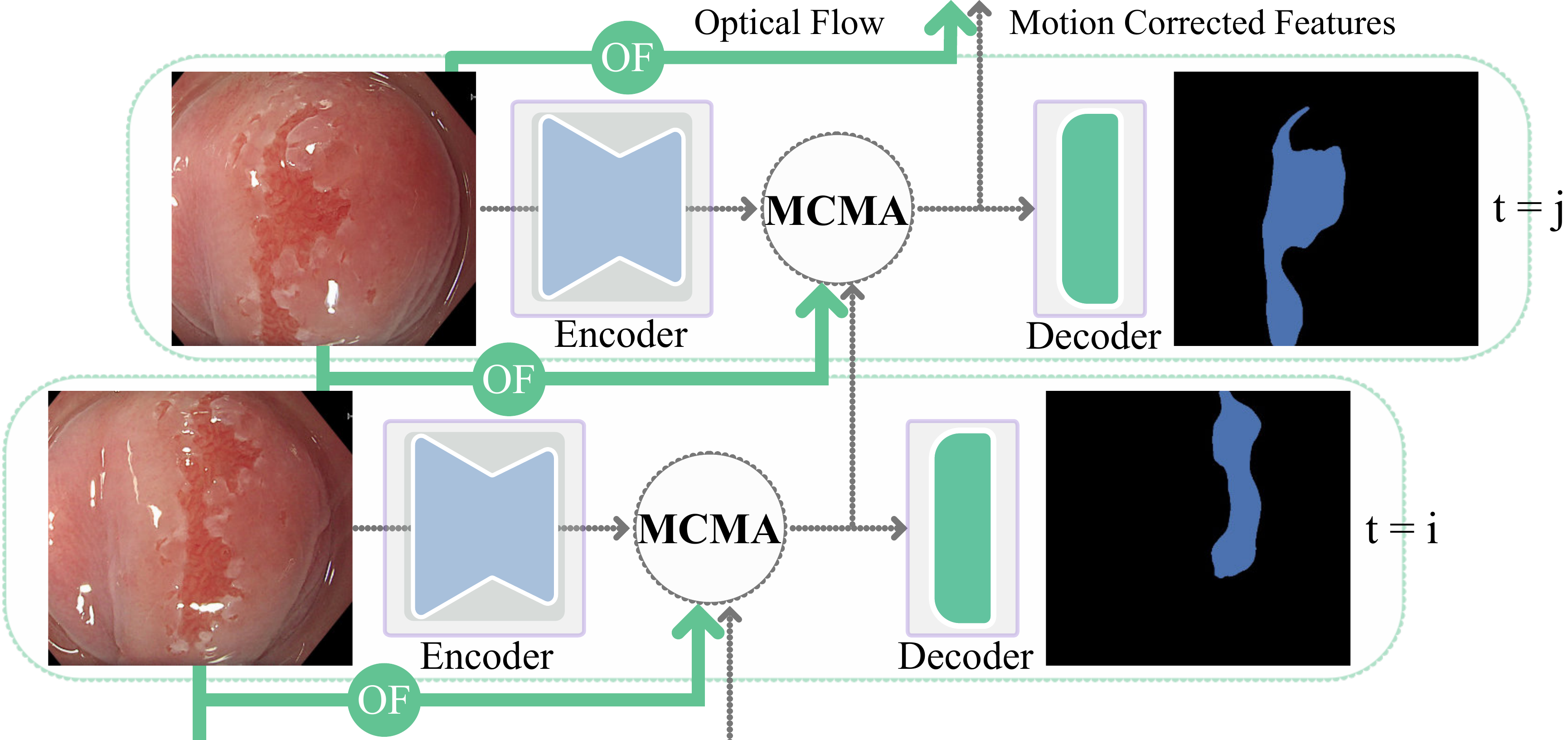}
\caption{
Inference with MCMA.
Features extracted by the encoder on the current frame, the estimated displacement field, and the results of the previous iteration are merged in the MCMA module.
The result is passed to the decoder layer to predict the segmentation and as input for the MCMA calculation in the next frame.}
\label{fig:mcma}
\end{figure*}
\subsection{Medical Video Semantic Segmentation}

In the medical domain, the focus of video semantic segmentation is often on improving the segmentation quality by incorporating temporal context into the network architecture \cite{attia2017surgical, lin2021multi, yang2022drr, li2021preserving, zhao2020learning, islam2021st, wang2021efficient}.
One approach is to use recurrent modules placed at the end of the encoder branch to aggregate the resulting low-level features \cite{attia2017surgical, lin2021multi} or to perform multi-frame feature aggregation in both the encoder and decoder branches at different stages \cite{yang2022drr}.
A related method is to employ LSTM modules in the network structure to temporally aggregate features, which can be done in the bottleneck of an encoder-decoder architecture \cite{li2021preserving, zhao2020learning} or in the decoder branch of such a network \cite{islam2021st}.
Another approach is based on ConvLSTM blocks as part of efficient local temporal aggregation and active global temporal aggregation modules to leverage both local temporal dependence and global semantic information \cite{wang2021efficient}.

In addition to these architecture-oriented approaches, other work uses optical flow between two consecutive video frames \cite{lin2019automatic, sestini2023fun, li2021preserving, zhao2020learning}.
In \cite{lin2019automatic}, the authors describe a multistep method that uses motion flow to determine the tip of a surgical instrument in the next frame.
A multistep method is also presented in \cite{sestini2023fun}, which consists of an unsupervised optic-flow segmentation and a generative-adversarial step for mapping between two domains.
Other works use optical flow-based warping to interpolate unlabeled intermediate frames in sparsely annotated video sequences \cite{li2021preserving} or in semi-supervised based approaches consisting of two branches to perform the recovery of the original annotation distribution and the compensation of fast instrument movements simultaneously \cite{zhao2020learning}.

\section{Motion-Corrected Moving Average}

Motion-Corrected Moving Average (MCMA) intends to include temporal information within the model without requiring alterations to the segmentation architecture that must be considered during training.
To apply MCMA to a video sequence of images $ \{\text{\textbf{x}}_1, \text{\textbf{x}}_2, ... \}$ the trained segmentation model needs to be partitioned into feature encoding and decoding subnetworks $E(\cdot)$ and $D(\cdot)$.
The encoder-decoder terminology does not have to refer to the architectural choices of the neural network.
The partitioning point for the encoder-decoder split in MCMA can be after the very first layers or at the penultimate layer of the model.
On an RGB input image $\text{\textbf{x}}_i \in \mathbb{R}^{1 \times 3 \times H \times W}$  the encoder network predicts a feature representation $\bm{f}_i = E(\text{\textbf{x}}_i)$  of dimension $\bm{f}_i \in \mathbb{R} ^ {1\times C \times h \times w}$, where the feature depth $C$ and resolution $h$ and $w$ depends on the partition point in the architecture.
Without temporal information $\bm{y}_i = D(\bm{f}_i)$ corresponds to the segmentation $\bm{y}_i \in \mathbb{R}^{1 \times L \times H \times W}$ for the $L$ classes.

\subsection{Motion Alignment in Feature Space}
Given two consecutive input images $\text{\textbf{x}}_i$ and $\text{\textbf{x}}_j$ the pixel displacement going from $i$ to $j$ can be estimated with an optical flow algorithm \cite{farne,flownetv2,siftflow}.
Similar to \cite{flowguidedobject}, we warp intermediate features by applying the estimated optical flow, denoted by $\mathcal{F}_{i \rightarrow j}$, bilinearly-interpolated to the feature resolution $h \times w$.
In practice, the estimated flow can exhibit inaccuracies and either under or overemphasize the movements, depending on the dataset and algorithm.
We introduce the scaling parameter $\lambda$ that is applied to the flow during the bilinear warping function $\mathcal{W}$ when adjusting the features:
\begin{equation}
    \phi_{\bm{f}_{i \rightarrow j}} := \mathcal{W}(\bm{f}_{i}, \lambda \mathcal{F}_{i \rightarrow j}).
\end{equation}

\begin{figure*}[t]

	\subfloat[]{ 
		\includegraphics[width=0.24\textwidth]{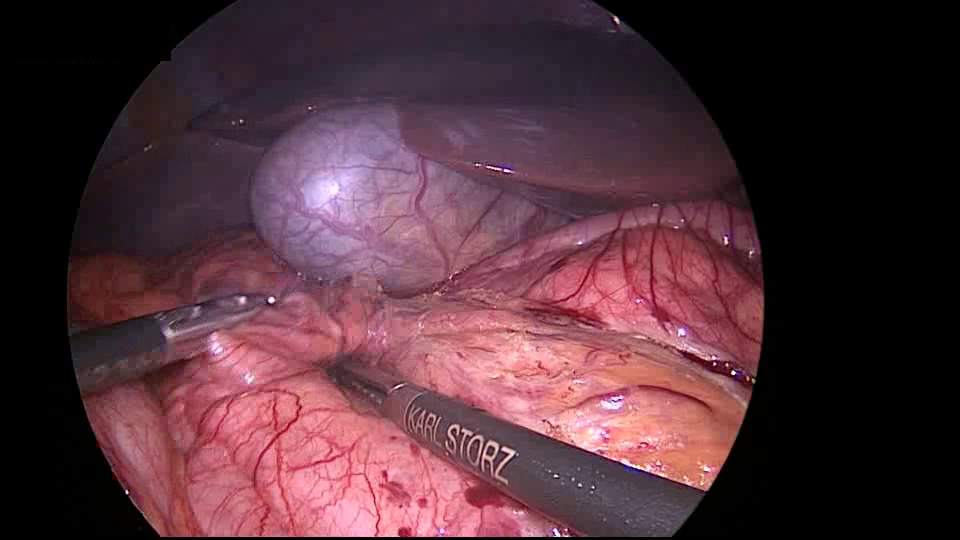}
        \label{datasets:ev_images}
    }
    \hfill
    \subfloat[]{
      	\includegraphics[width=0.24\textwidth]{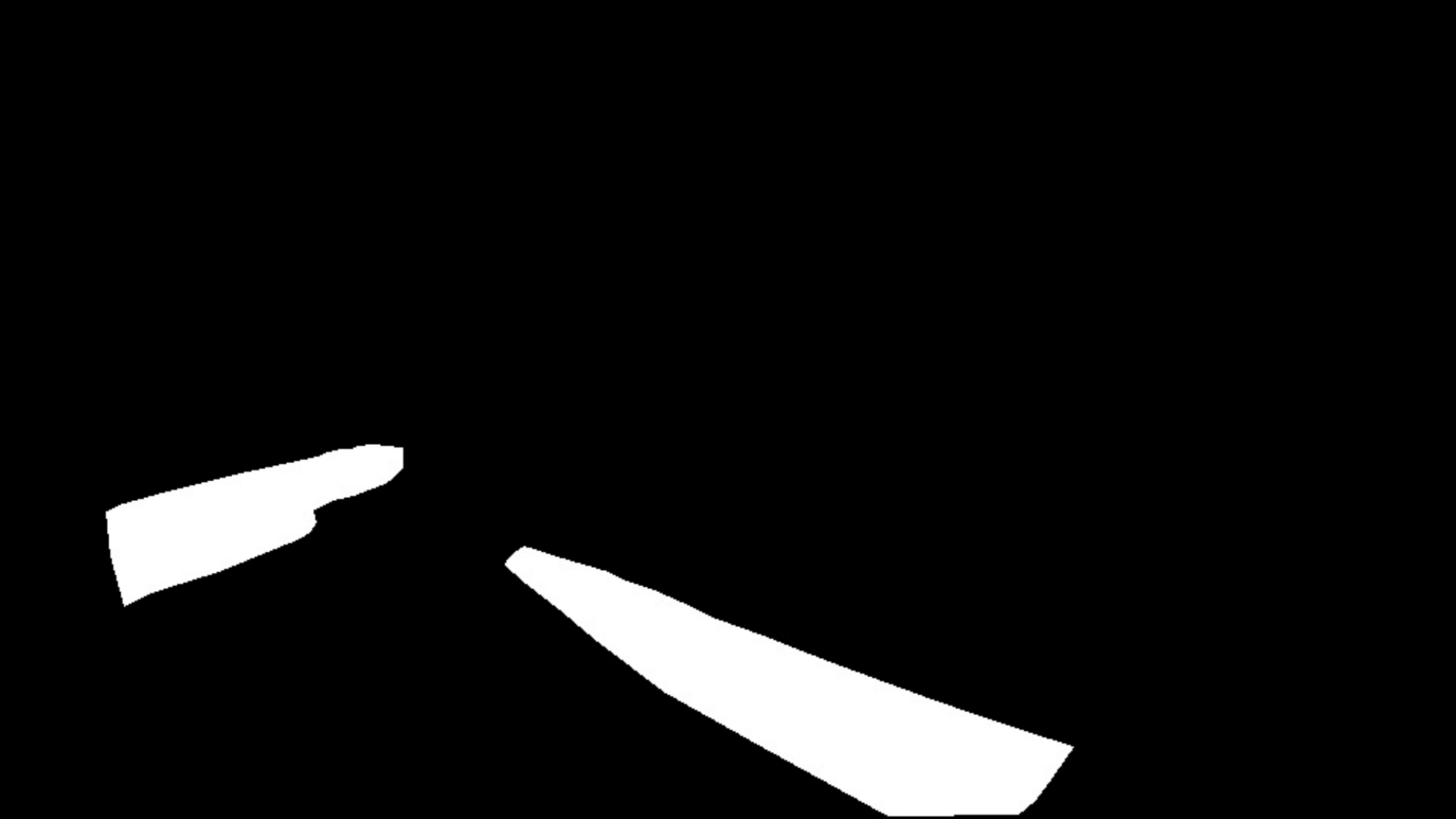}
       \label{datasets:ev_mask}   
    }
    \hfill
    \subfloat[]{
      	\includegraphics[width=0.24\textwidth]{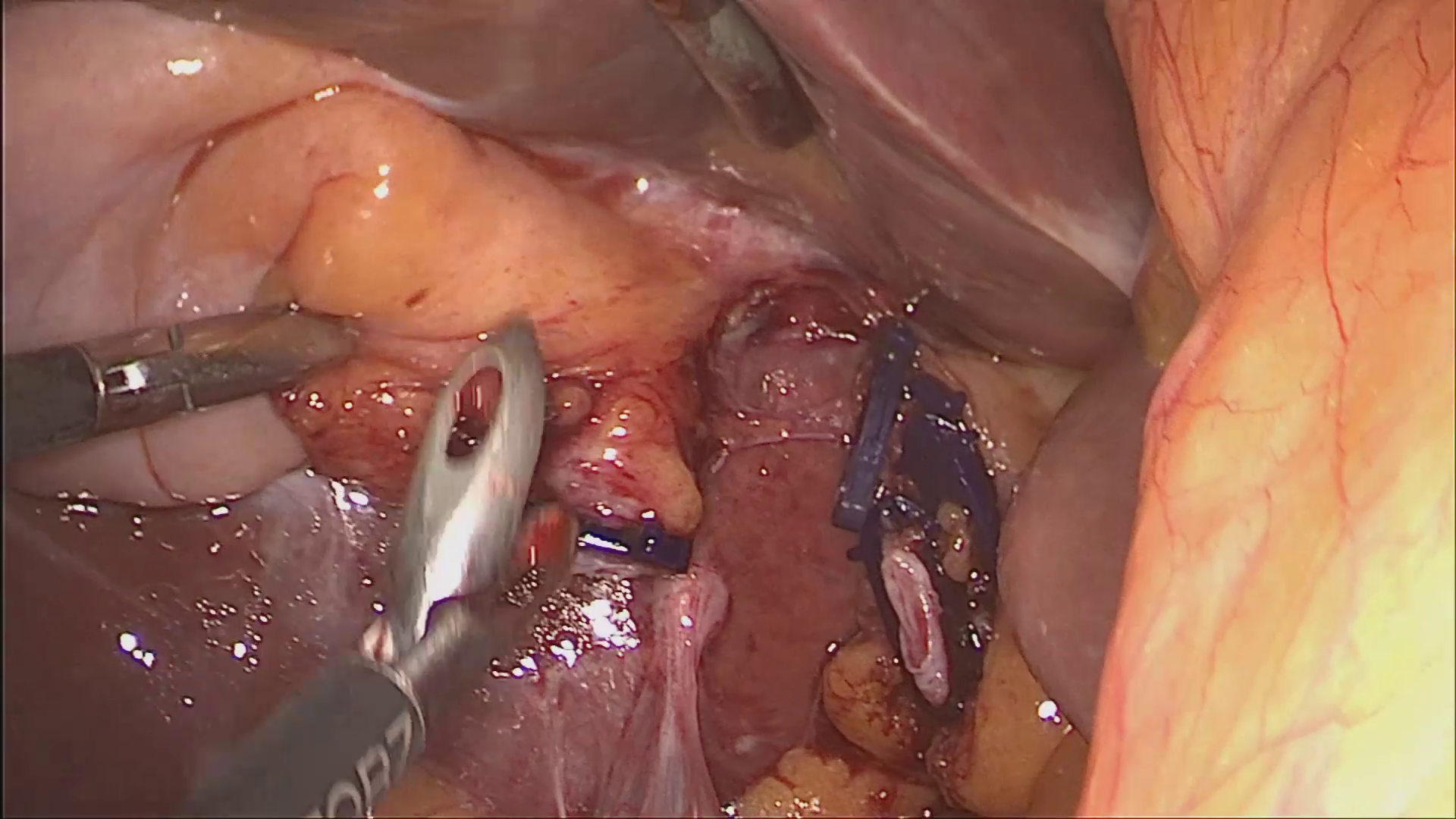}
        \label{datasets:cholec_images}
    }
    \hfill
    \subfloat[]{
      	\includegraphics[width=0.24\textwidth]{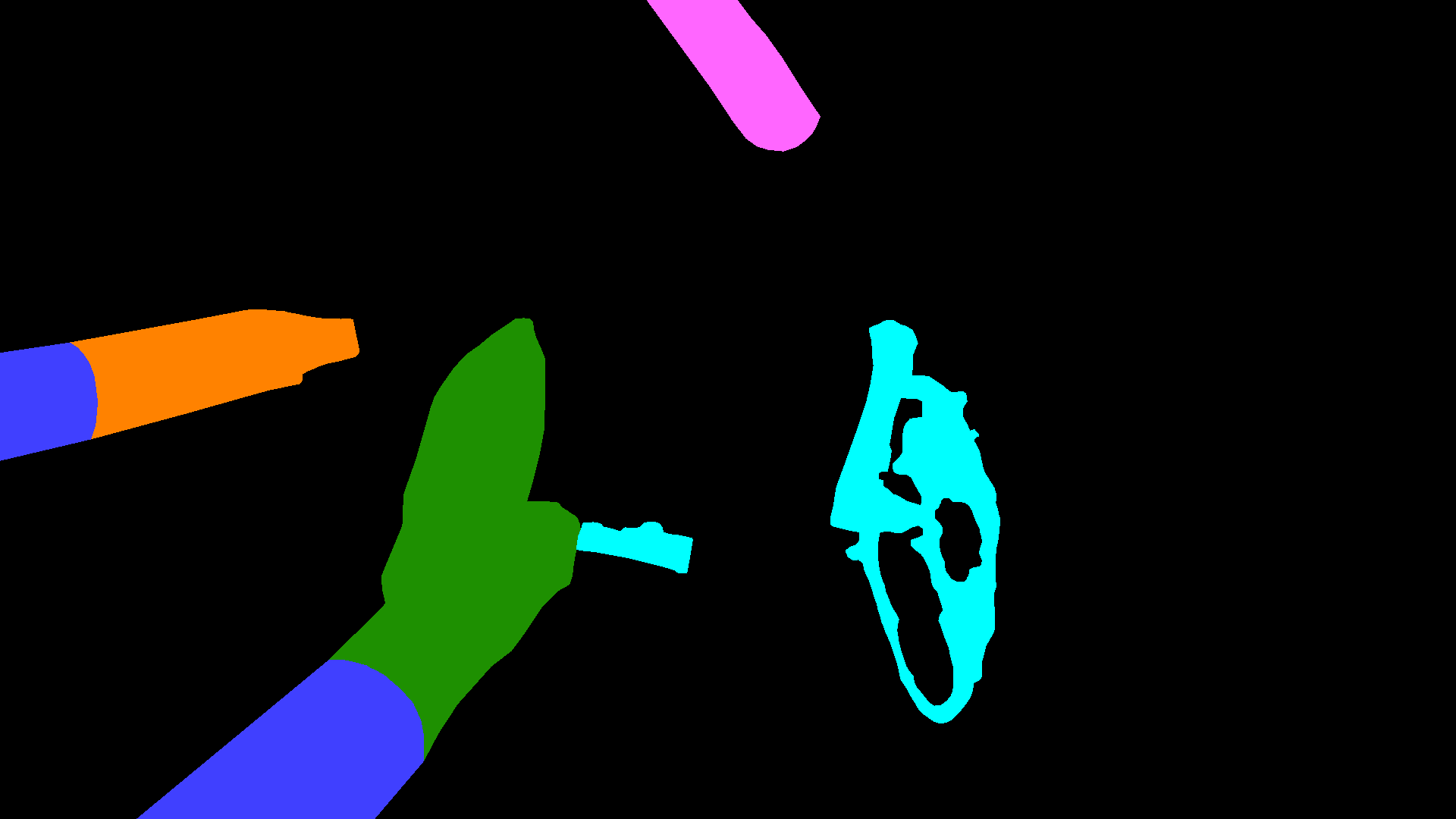}
        \label{datasets:cholec_mask}
    }
    \caption{Example source images with corresponding segmentation masks from the EndoVis-2019 \cite{ross2021comparative} and Cholec datasets.}
    \label{datasets:example_images}
\end{figure*}
\subsection{Feature Space Exponential Moving Average}
With the now available warped features, the question arises of how past and present should be merged without leading to architecture adjustments.

An elementary approach for incorporating temporal information is to include past outputs, either complete segmentation or intermediary features, in a moving average calculation.
Consequently, the motion correction can correct the spatial disconnect a low $\alpha$ can cause in the moving average calculation.
For the current input $\text{\textbf{x}}_j$ the motion-corrected moving average $\bm{f}^{\prime}_j$ is calculated as a combination of the current features, flow, and $\bm{f}^{\prime}_i$ from the previous step:
\begin{equation}
    \bm{f}^{\prime}_j = \alpha \bm{f}_j + (1-\alpha)\phi_{\bm{f^{\prime}}_{i \rightarrow j}}.
    \label{eq:mcma}
\end{equation}

The augmented features are then used as inputs for the decoder network and produce the adjusted segmentation $\bm{y}^{\prime}_{j} = D(\bm{f}^{\prime}_j)$.

Figure \ref{fig:mcma} illustrates the iterative processing at inference and how the components interact.

\section{Datasets and Setup}
\label{sec:setup}
All experiments use a pre-trained DeepLabv3+ \cite{deeplab} with a ResNet \cite{resnet} backbone, implemented in PyTorch \cite{pytorch}.
The scaling parameter $\lambda$ is set to $2.0$ for all experiments.
On Barrett and Cityscapes, the model with a ResNet-50 backbone was trained for 120 and 90 epochs, respectively, with SGD, and a learning rate of 0.1 with polynomial decay.
On EndoVis-2019 and Cholec, the ResNet-101-based models were trained for a total of 60 epochs each, with Adam, an initial learning rate of $1e-4$, and with cyclic learning rate decay.

For MCMA, we chose the encoder-decoder split after the low-level and upsampled features are concatenated in the DeepLabv3+ architecture. 
The optical flow is implemented with the cuda accelerated Farneback algorithm \cite{farne}, and the Nvidia Optical Flow SDK 2.0, both provided by OpenCV \cite{opencv}.

\subsection{Barrett}
The proprietary dataset consists of images and videos of the upper gastrointestinal tract. 
To focus is the differentiation between Barrett's esophagus (BE) and Barrett's esophagus-related neoplasia (BERN). 
The training split of the dataset consists of 3080 images where an image-level binary diagnosis and a three-class segmentation between background, BE, and BERN are available.
A subset of the data is available in up to four imaging modalities: white light and narrow-band imaging with and without vinegar.
No fully sequentially labeled data was available to train the neural network. 
The Barrett results are calculated on video cases unseen during training.
Of the 17 50\,Hz video sequences ranging from 30 to 90 seconds, nine exclusively show non-dysplastic Barrett, and eight contain a neoplastic lesion.
All data is histologically confirmed and labeled by experienced gastroenterologists.
By visual inspection,  at least one sequence per video was selected, where at least two classes were present, and a large amount of movement was observable (see the samples in Fig. \ref{fig:comparison}).
All frames are resized to a resolution of $640 \times 512 $ pixels.
At least one frame is annotated for the selected sequences, resulting in 180 labeled samples of the 17 videos.
To measure the effects of EMA and MCMA, we iteratively apply the algorithm to the 100 frames preceding the labeled frame.
All frames of the BE videos are utilized to evaluate the pixel-wise false positive rate.

\subsection{EndoVis-2019}

The robust medical instrument segmentation \cite{ross2021comparative} dataset was part of the 2019 MICCAI Endoscopic Vision (EndoVis) sub-challenge. 
The dataset includes video recordings of 30 surgical procedures divided evenly into three types.
The sequences show in-vivo human surgeries performed with rigid instruments, namely rectal resection procedures, proctocolectomy procedures, and sigmoid resection procedures.
All videos were recorded at 25\,Hz and provide binary annotations indicating the surgical instrument or background on 10,040 frames (Fig. \ref{datasets:ev_images}, \ref{datasets:ev_mask}).
The evaluation tasks are divided into three stages of increasing complexity.
In the first stage, the test dataset is taken from the same procedure as the training data. 
Stage two uses sequences of new, unseen patients for testing.
In stage three, the surgical procedures were recorded from both unseen patients and a different type of surgery.

To evaluate MCMA, we follow the same structure as the organizers of the sub-challenge.
To determine the effects of EMA and MCMA, we iteratively apply the algorithm to 250 unlabeled frames that precede an annotated frame.

\subsection{Cholec}

The proprietary Cholec dataset consists of eight video sequences showing real-life minimally invasive cholecystectomies, with durations ranging from 23 to 60 minutes.
Six recordings are captured at 25\,Hz and the remaining two at 50\,Hz.
In each sequence, a ground truth segmentation mask is provided for one frame per second, in which each pixel is assigned to one of 18 classes (Fig. \ref{datasets:cholec_images}, \ref{datasets:cholec_mask}).

In order to capture the influence of EMA and the proposed MCMA method, the algorithm is iteratively executed for all images of a video sequence and evaluated on the annotated frames.
To obtain statistically meaningful results, we perform a two-fold cross-validation five times \cite{dietterich1998approximate} and calculate the arithmetic mean of the results.

\subsection{Cityscapes}
Although this paper focuses on medical imaging, our method is not restricted to this domain.
It can be applied to other semantic segmentation datasets - like the Cityscapes \cite{Cordts2016Cityscapes} dataset.
The Cityscapes dataset spans 2975 labeled images and 500 validation images with a resolution of $2048 \times 1024 $. 
The benchmarking task consists of 19 classes, and the data is densely labeled.
Part of the extended dataset are the unlabeled 30\,Hz sequences showing the 19 preceding and ten frames following each labeled training and validation sample.
We report the validation set performance and iteratively include the 19 preceding frames for the EMA and MCMA calculations.

 \begin{table*}[t]
	\centering
	    \caption{Results on the used datasets partitioned to show the mIoU (in $\%$) on the whole dataset ($100\%$) and only on the frames with the most ($\uparrow 20\%$), least ($\downarrow 20\%$) and intermediate ($60\%$) amount of movements.}
	    \label{tab:results}

      \begin{tabular*}{\textwidth}{@{\extracolsep{\fill}} l rrrr rrrr }
                \toprule
                
                \multicolumn{9}{c}{\textbf{Proprietary Datasets}}
                \\
                \cmidrule{2-5}
                \cmidrule{6-9}
                
                &
                \multicolumn{4}{c}{Barrett} & 
                \multicolumn{4}{c}{Cholec} \\

                \cmidrule{2-5}
                \cmidrule{6-9}

                Method &
                \multicolumn{1}{c}{100$\%$} &
                \multicolumn{1}{c}{$\uparrow 20\%$} & 
                \multicolumn{1}{c}{$\downarrow 20\%$} & 
                \multicolumn{1}{c}{60$\%$} & 
                \multicolumn{1}{c}{100$\%$} & 
                \multicolumn{1}{c}{$\uparrow 20\%$} & 
                \multicolumn{1}{c}{$\downarrow 20\%$} & 
                \multicolumn{1}{c}{60$\%$} \\

                \cmidrule{2-5}
                \cmidrule{6-9}

                Baseline & 76.45 & 73.10 & 76.41 & 77.04 & 
                38.19 & 40.54 & 26.28 & 41.09 \\      
                EMA & 69.58 & 61.17 & 73.67 & 70.42 & 
                38.31 & 40.70 & 26.42 & 41.24 \\  

                MCMA & \textbf{79.13} & \textbf{75.86} & 
                \textbf{76.52} & \textbf{80.85} &
                \textbf{38.50} & \textbf{40.91} & 
                \textbf{26.67} & \textbf{41.45}\\    

                \cmidrule{2-5}
                \cmidrule{6-9}
                
                \multicolumn{9}{c}{\textbf{Public Datasets}}
                \\
                \cmidrule{2-5}
                \cmidrule{6-9}
                &
                \multicolumn{4}{c}{EndoVis-2019 - Stage 1 \cite{ross2021comparative}} & 
                \multicolumn{4}{c}{EndoVis-2019 - Stage 2 \cite{ross2021comparative}} \\

                \cmidrule{2-5}
                \cmidrule{6-9}

                &
                \multicolumn{1}{c}{100$\%$} &
                \multicolumn{1}{c}{$\uparrow 20\%$} & 
                \multicolumn{1}{c}{$\downarrow 20\%$} & 
                \multicolumn{1}{c}{60$\%$} & 
                \multicolumn{1}{c}{100$\%$} & 
                \multicolumn{1}{c}{$\uparrow 20\%$} & 
                \multicolumn{1}{c}{$\downarrow 20\%$} & 
                \multicolumn{1}{c}{60$\%$} \\

                \cmidrule{2-5}
                \cmidrule{6-9}

                Baseline & 88.30 & \textbf{92.05} & 86.33 & 86.37 & 
                79.80 & 71.00 & \textbf{77.10} & \textbf{85.56} \\      
                EMA & 88.34 & 85.42 & 93.18 & 87.88 & 
                79.80 & 81.09 & 65.55 & 83.90 \\  

                MCMA & \textbf{88.45} & 85.66 & 
                \textbf{93.21} & \textbf{87.96} &
                \textbf{79.82} & \textbf{81.12} &
                65.56 & 83.92 \\

                \cmidrule{2-5}
                \cmidrule{6-9}
                &
                \multicolumn{4}{c}{EndoVis-2019 - Stage 3 \cite{ross2021comparative}} & 
                \multicolumn{4}{c}{Cityscapes \cite{Cordts2016Cityscapes}} \\

                \cmidrule{2-5}
                \cmidrule{6-9}

                &
                \multicolumn{1}{c}{100$\%$} &
                \multicolumn{1}{c}{$\uparrow 20\%$} & 
                \multicolumn{1}{c}{$\downarrow 20\%$} & 
                \multicolumn{1}{c}{60$\%$} & 
                \multicolumn{1}{c}{100$\%$} & 
                \multicolumn{1}{c}{$\uparrow 20\%$} & 
                \multicolumn{1}{c}{$\downarrow 20\%$} & 
                \multicolumn{1}{c}{60$\%$} \\

                \cmidrule{2-5}
                \cmidrule{6-9}

                Baseline & 75.47 & 72.39 & \textbf{72.89} & \textbf{77.68} & 
                78.19 & 75.18 & 76.67 & 78.19 \\      
                EMA & 75.45 & 75.92 & 68.91 & 77.45 & 
                76.83 & 72.95 & 76.12 & 76.78 \\  

                MCMA & \textbf{75.50} & \textbf{75.95} &
                68.88 & 77.54 & 
                \textbf{78.51} & \textbf{75.61} & 
                \textbf{76.73} & \textbf{78.64} \\    
                
                \bottomrule
	    \end{tabular*}
    \end{table*}
\section{Results and Discussion}

On all datasets, we report the mean Intersection over Union for the labeled frames and partitioned to the $80$\si{\percent} and $20$\si{\percent} quantiles and the remaining $60$\si{\percent} of the data.
The quantiles are calculated for the mean length of the displacement vectors in the optical flow and thus help to evaluate how the algorithm performs on the subset of frames with large, small, and intermediate amounts of motion.
For Barrett's, we provide further hyperparameter analysis and discuss the runtime performance of MCMA in sequential and parallel execution settings.

\subsection{Barrett}
\begin{figure*}[t]
\centering
\includegraphics[width=1.0\textwidth]{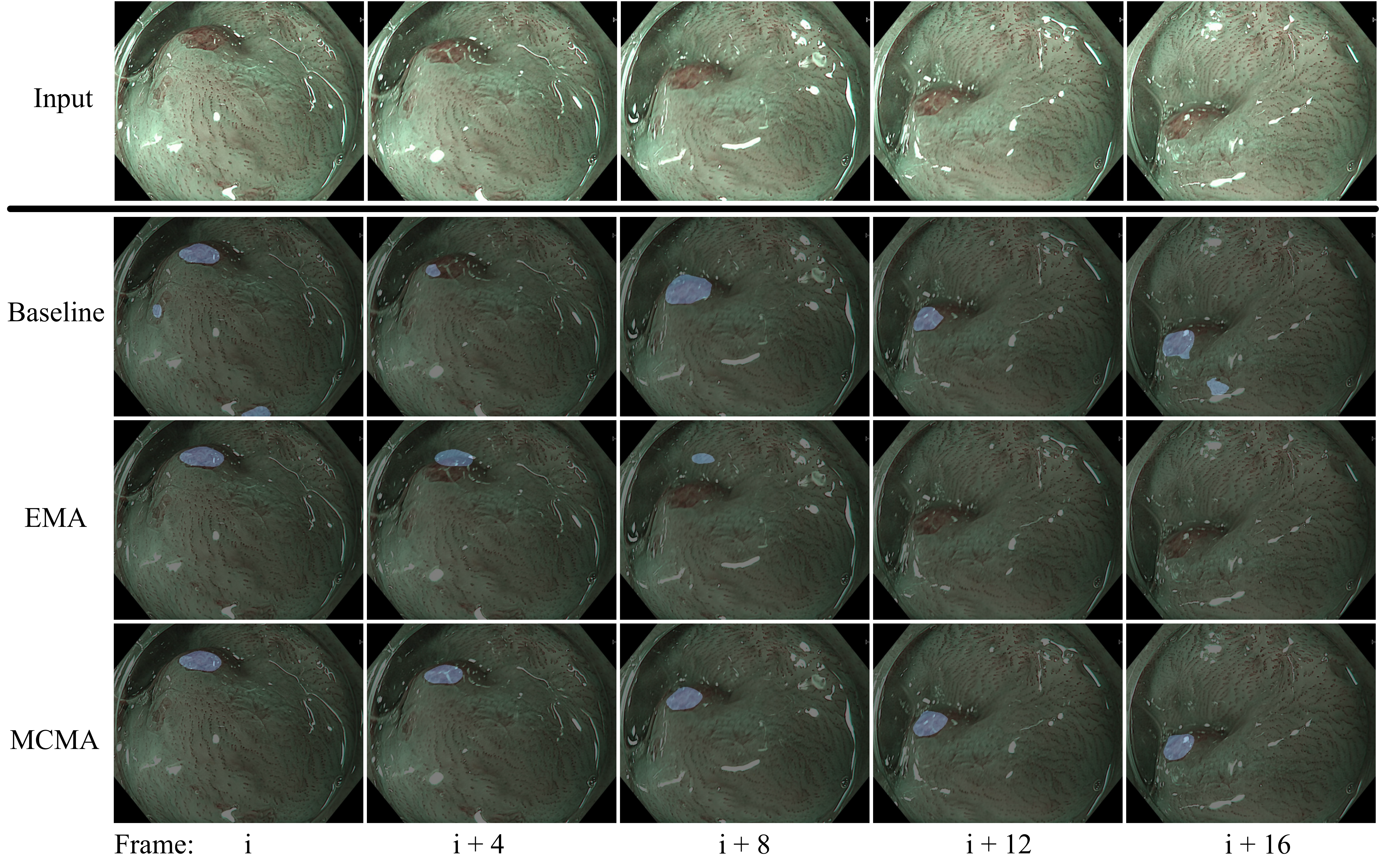}
\caption{
Comparison between the baseline, EMA, and MCMA on consecutive frames of the the Barretts dataset.
The image-based segmentation can track the non-dysplastic Barrets (blue) frame by frame but introduces visual noise.
With a low $\alpha$ value, EMA loses the segmentation during fast movements.
MCMA, on the other hand, suppresses visual noise while accurately tracking the relevant region.
}
\label{fig:comparison}
\end{figure*}
\label{result:barrett}
MCMA proves to be the most beneficial on the Barrett dataset.
The first section in Table \ref{tab:results} shows the performance of MCMA compared to just EMA and single image segmentation.
The results were obtained with an $\alpha$ of $0.1$, a very slow-moving average value. 
The difference between EMA and the other approaches signifies that fast movements and slow averages significantly deteriorate performance. 
An example is given in Figure~\ref{fig:comparison}.
Because of the slow average, EMA first introduces ghosting and then loses the segmentation of the Barrett's esophagus region.
Both the baseline and MCMA keep track of the relevant region.

\subsubsection{Suppressing False Positives}
\begin{table}[b]
	\begin{center}
	    \caption{Rate of false positive BERN detection on the nine cases showing no neoplastic lesions.
}
	    \label{tab:results-falsepos}
	    \setlength{\tabcolsep}{4pt}
	    \begin{tabular}{@{} l | c c  @{} }

		 & on all frames & on $FP$ frames \\
  \toprule
		Baseline & 1.11\si{\percent} & 4.13\si{\percent} \\
		EMA & 0.96\si{\percent} & 3.37\si{\percent} \\
\rowcolor{Gray} MCMA & \textbf{0.77\si{\percent}} & \textbf{2.78}\si{\percent} \\
	    \end{tabular}
	\end{center}
    \end{table}
The slow-moving average also reduces visual noise and false positives and suppresses the inter-frame prediction variability.
Comparing the area covered by pixels predicted as BERN for the nine video cases with no neoplasia present, both averaging approaches reduce the relative number of false positive ($FP$) pixels, shown in Table \ref{tab:results-falsepos}. Just EMA reduces the $FP$ rate from $1.11$\si{\percent} to $0.96$\si{\percent} and MCMA further decreases the rate to $0.77$\si{\percent}.
Focusing only on the subset where the baseline model falsely identified BERN regions, MCMA significantly reduces the rate from $4.13$\si{\percent} to $2.78$\si{\percent}.
Coupled with Table \ref{tab:results}, these results strengthen the position of MCMA as it leads to shape-accurate predictions while also reducing visual noise during video processing.

\subsubsection{Changing the Smoothing Factor}
\begin{figure*}[t]
\centering
\includegraphics[width=0.9\textwidth]{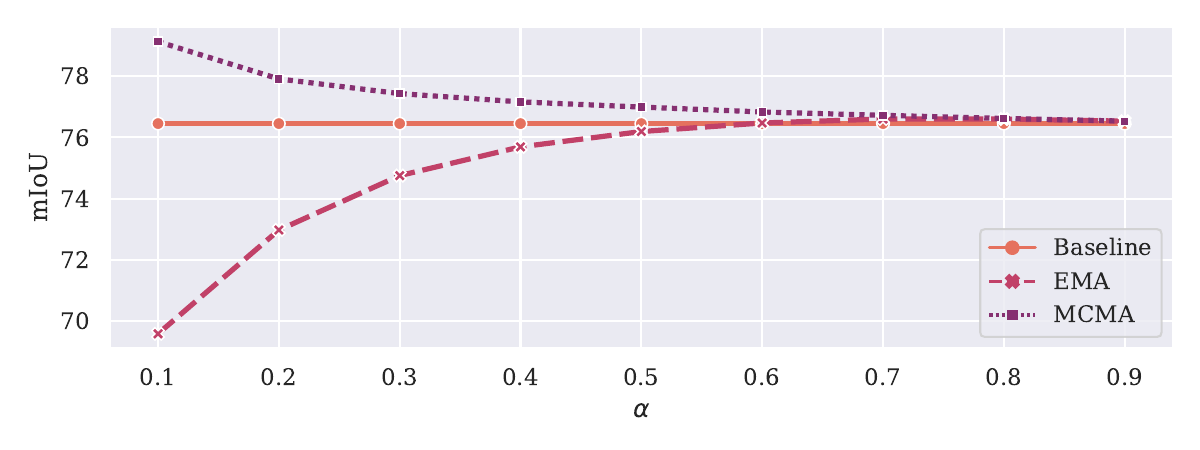}
\caption{Effects of changing $\alpha$ for EMA and MCMA on the Barrett dataset. With a slow moving average, MCMA can achieve the most accurate results, while the quality with just EMA deteriorates. The more focus is put on the recent frames, the smaller the difference between all three approaches becomes.}
\label{fig:barrettalpha}
\end{figure*}

The size of $\alpha$ determines how much emphasis is put on previous results in the exponential moving average.
As shown in section \ref{result:barrett}, on Barrett, MCMA achieves considerable improvements over EMA and the baseline with a very slow-moving average.
Figure \ref{fig:barrettalpha} extends these results and highlights the effects of changing the $\alpha$ value between $0.1$ and $0.9$.
MCMA consistently achieves the highest mIoU; the faster the moving average becomes, the less pronounced the differences are.

\subsubsection{Runtime Performance}
\begin{figure}[t]
\centering
    \includegraphics[width=0.9\linewidth]{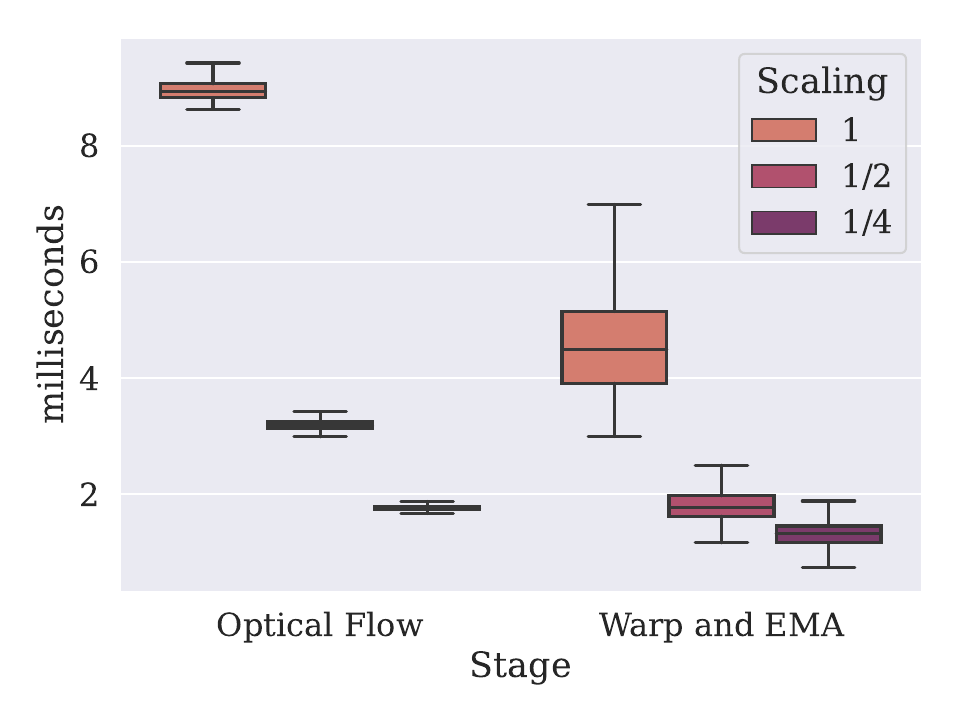}
    \caption{
    Runtime performance of MCMA, segmented into the optical flow component and the warping and EMA calculation (eq. (\ref{eq:mcma})) for varying resolution scales. The computation time for both parts benefits from a lower resolution. Results were obtained on an Nvidia RTX 3090.
    }
    \label{fig:seqperf}

\end{figure}

Apart from the accuracy of the predictions, the computational overhead MCMA introduces is equally important.
MCMA consists of two expensive operations: a neural network's forward pass and dense optical flow calculation.
Still, modern hardware and implementations lessen the runtime burden and enable real-time computations.
For the optical flow, we have opted for a cuda-supported implementation (sec. \ref{sec:setup}).
One user-definable parameter that considerably affects execution time is the input resolution for the optical flow calculation.
We demonstrate the effects of keeping the initial resolution and compare it with scalings of $1/2$ and $1/4$, resulting in flow maps of $320\times 256$ and $160\times128$ width and height.
Figure \ref{fig:seqperf} gives an overview of the runtime cost of both the dense optical flow and the bilinear warping and feature-space EMA calculation on the Barrett's validation data.
Reducing the resolution decreases the computational cost significantly, from $13.83~(\pm 1.83)$ milliseconds to $3.21~(\pm 0.83)$ milliseconds, on a Nvidia RTX 3090 with the Nvidia Optical Flow 2.0 SDK, while not affecting the accuracy of MCMA (Table \ref{tab:barrett-perf}). 
\begin{figure}[t]

\centering
    \includegraphics[width=0.9\linewidth]{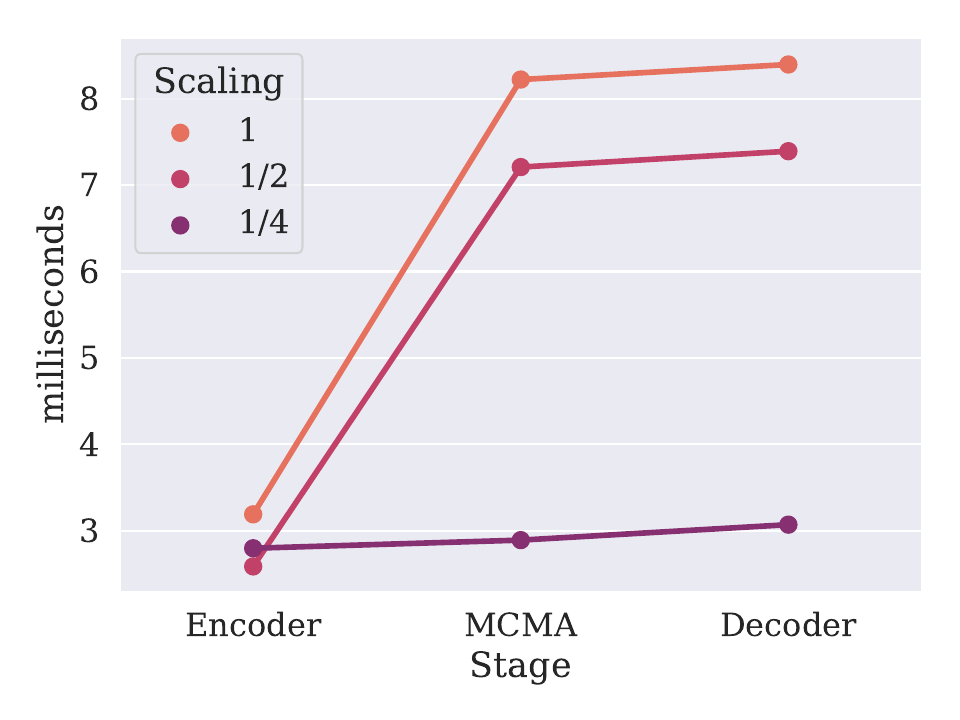}
    \caption{
 Forward pass duration with a parallel MCMA implementation at varying resolution scales on an Nvidia RTX 3090. Both the encoding forward pass and the optical flow calculation can be done in parallel. At small resolution scales, this leads to the optical flow calculation and MCMA as a whole not causing considerable overhead.
    }
    \label{fig:parperf}
\end{figure}
Further, Figure \ref{fig:parperf} shows the execution times of MCMA over a $\sim$1-minute 50\,Hz video sequence with a multi-threaded architecture implemented in C++.
The optical flow calculation and encoding forward pass are run in parallel on a single Nvidia RTX 3090.
At a low-resolution scale, the parallel execution effectively removes the time constraint of the optical flow.
The warping is fast and equally benefits from the smaller resolution, only taking $0.09~(\pm 0.12)$ milliseconds.

On average, the complete forward pass with MCMA takes $3.07~(\pm0.39)$ milliseconds with the multi-threaded architecture, potentially enabling frequencies of $>$ 150\,Hz at the given image resolution of $640 \times 512$ and optical flow resolution scale of $1/4$.

\begin{table}[t]
	\begin{center}
     \caption{Performance on the Barrett dataset with varying resolution scaling. The optical flow algorithm effectively achieves identical performance, independent of the resolution.} 
	    \label{tab:barrett-perf}
	    
	    \setlength{\tabcolsep}{10pt}
     
	    \begin{tabularx}{\textwidth}{ l r | c c c }
        & & & Scaling & \\
		        Dataset & Proportion &  $1/4$ &  $1/2$  & $1$   \\
            \cmidrule{1-5}
		\multirow{4}{*}{Barrett} &$100\%$ & 79.11 & 79.11 & \textbf{79.13} \\
	       & $\uparrow 20\%$ & 75.37 & 75.38 & \textbf{75.86} \\
             & $\downarrow 20\%$ & \textbf{76.52} & \textbf{76.52} & \textbf{76.52} \\
                        & $60\%$ & 80.96 & \textbf{80.97} & 80.85 \\

	    \end{tabularx}
     \end{center}
    \end{table}

Even at a full resolution scale, a parallel implementation still results in an average execution time of $8.39~(\pm0.47)$ milliseconds, far below the 20 milliseconds available per frame for smooth 50\,Hz execution.

\subsection{EndoVis-2019}

The results for the three stages of the EndoVis-2019 dataset are shown in the second, third, and fourth sections of Table~\ref{tab:results}.
The $\alpha$ values used for the EMA and MCMA experiments were $0.65$, $0.95$, and $0.90$, respectively.
Regarding stage one, MCMA achieved the best results in three of the four categories, except for the 20\si{\percent} frames with the highest motion, where the baseline method achieved the best results.
For stages two and three, MCMA was the best approach on the entire dataset and regarding the 20\si{\percent} frames with the most motion, whereas the results for slow and moderate motion video frames were superior using the baseline method.
The high segmentation quality baseline especially concerning slow and moderate moving images could result from the task's simplicity and shows a small variance between the training and test data, allowing the network to perform accurately without temporal information.
The choice of high alpha values compared to the other datasets also indicates that the inclusion of previous predictions has little impact on the quality of the segmentation of the current image. 
Furthermore, the assignment of video frames to the respective motion categories is based on the overall image movement, i.e., the average length of the displacement vector of all pixels although the relevant regions, represented by the surgical instruments, often represent only a relatively small part of the image.
This characteristic distinguishes the EndoVis-2019 dataset from the other datasets in which the relevant areas cover a larger portion of the entire frame, either by the nature of the data and the underlying task (Barrett, Cityscapes) or by a closer inclusion of the surgical tools along with more classes annotated in the images such as clips for hemostasis or retrieval bags, which temporarily may also cover a larger area (Cholec).
MCMA leads to a significantly better mean IoU in the third stage and improved results for the 20\si{\percent} of video frames that contain the most amount of movement.
The third stage represents the most challenging task, and here, MCMA can demonstrate the benefit of including temporal information.

\subsection{Cholec}
\label{sec:cholec}
The results averaged over all ten folds for the Cholec dataset were obtained with an $\alpha$ of $0.75$ and are shown in the fifth section of Table~\ref{tab:results}.
The proposed MCMA method leads to better mean IoU than the single image segmentation and plain EMA.
The largest deviations from the baseline are on the frames with the most and least amount of motion.
The results suggest that the temporal information suppresses outliers in stable parts of the video while also accurately warping the features when movement is present.

\subsection{Cityscapes}
Cityscapes differs in that it is densely labeled without an obvious distinction between foreground and background.
Here MCMA achieves the best results for $\alpha = 0.65$, shown in Table \ref{tab:results}.
Compared to the performance in Section \ref{sec:cholec} the difference to the baseline grows with the amount of motion.
The result suggests a low variance in the slow-moving sequences but difficulties in faster scenes, which MCMA can help to amend.

\section{Conclusion}
Motion-Corrected Moving Average allows the inclusion of temporal information during inference while having a low computational footprint and no training requirements for the model or the dataset.
Our experiments have shown improvements with MCMA to varying degrees.
The chosen datasets range from binary foreground-background segmentation to densely labeled tasks, and the general gains achievable with MCMA demonstrate the versatility of the approach.

Although optical flow is potentially costly to compute, our approach allows a parallel execution on the GPU and demonstrably leads to significantly reduced runtime overhead.
Apart from the low overhead, not requiring specific training allows a wide range of applications.
Without this constraint, MCMA enables video segmentation applications where no labeled video sequences, even no video sequences, are available as training data or when the segmentation model is integrated into a multi-task architecture that does not allow architecture alterations required by other methods.

Determining the smoothing factor depending on the nature of the dataset used and the considered task represents a central opportunity for further work.
The dynamic, individual estimation of the smoothing factor for each pixel in a video frame also offers potential for improvement and further development of the method in the future.

\section{Acknowledgements}
\label{acknowledgements}

\noindent The authors thank the Augsburg University Hospital for curating and providing the
Barrett dataset.

\bibliographystyle{unsrtnat}

\bibliography{main}

\end{document}